\documentclass[paper]{ieice}
\usepackage{amsmath,amssymb,amsfonts} 
\usepackage{newtxtext}
\usepackage[varg]{newtxmath}

\usepackage{graphicx}
\usepackage{booktabs} 
\usepackage{multirow} 
\usepackage{threeparttable}
\usepackage{url}
\usepackage{flushend}
\usepackage{caption}

\field{}
\title{Robust Recommendation with Implicit Feedback via Eliminating the Effects of Unexpected Behaviors}
\authorlist{%
 \authorentry[gxuchenjie04@gmail.com]{Jie Chen}{membership}{1}\MembershipNumber{}
 \authorentry[wxxjlf@sina.com]{Lifen Jiang}{membership}{1}\MembershipNumber{}
 \authorentry[mcmxhd@163.com]{Chunmei Ma}{membership}{1}\MembershipNumber{}
 \authorentry[sunhuazhi@tjnu.edu.cn]{Huazhi Sun}{membership}{1}
}
\affiliate[1]{
	School of Computer and Information Engineering, Tianjin Normal University, Tianjin  300387, China
}

\received{2019}{8}{1}
\revised{2019}{8}{1}

\begin{document}
\maketitle
\begin{summary}
In the implicit feedback recommendation, incorporating short-term preference into recommender systems has attracted increasing attention in recent years. However, unexpected behaviors in historical interactions like clicking some items by accident don't well reflect users' inherent preferences. Existing studies fail to model the effects of unexpected behaviors thus achieve inferior recommendation performance.

In this paper, we propose a \textit{Multi-Preferences Model} (MPM) to eliminate the effects of unexpected behaviors. MPM first extracts the users' instant preferences from their recent historical interactions by a fine-grained preferences module. Then an unexpected-behaviors detector is trained to judge whether these instant preferences are biased by unexpected behaviors.  we also integrate user's general preference in MPM. Finally, an output module is performed to eliminates the effects of unexpected behaviors and integrates all the information to make a final recommendation. We conduct extensive experiments on two datasets of a movie and an e-retailing, demonstrating significant improvements in our model over the state-of-the-art methods. The experimental results show that MPM gets a massive improvement in HR@10 and NDCG@10, which relatively increased by 3.643\% and 4.107\% compare with AttRec model on average. We publish our code at https://github.com/chenjie04/MPM/.
\end{summary}

\section{Introduction}

Implicit feedback recommendation has attracted increasing attention in recent years as the convenience of data collection\cite{he2016fast} and it can take full advantage of users' interactions (such as view, click, purchase and so on) for recommendation. Prior efforts have shown the importance of exploiting users' behaviors for comprehending users' interest\cite{li2018learning,zhou2018micro,liu2018stamp:}. Existing works benefit from learning short-term preference by exploring users' recent historical interactions. As integrating the short-term preference, recommender systems have fully considered the evolution of users' preferences. 

\begin{figure}[!ht]
	\centering
	\vspace{-20pt}
	\setlength{\abovecaptionskip}{0pt}
	\setlength{\belowcaptionskip}{-10pt}
	\includegraphics[width=0.8\linewidth]{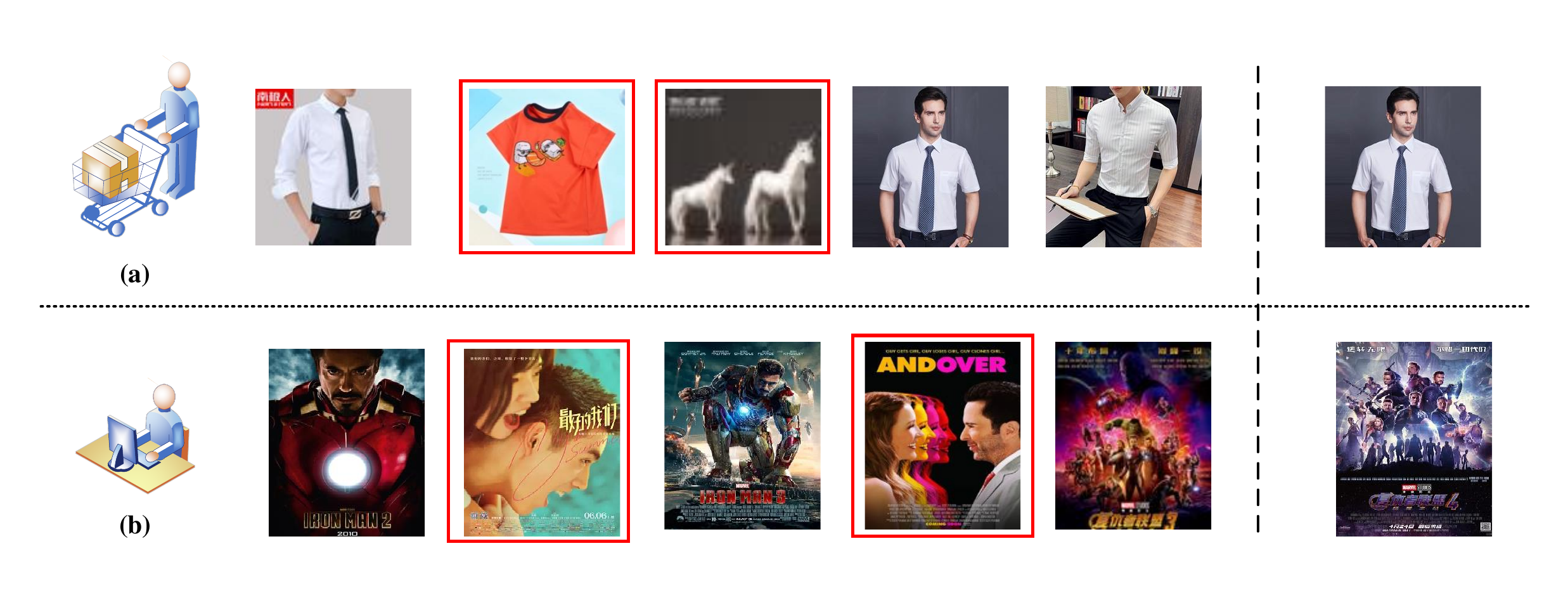}
	\caption{Unexpected behaviors in the historical interactions, as shown in red boxes.}
	\label{fig: unexpected_behaviors}
\end{figure}

However, in such an implicit feedback setting, a user's exact preference level is hard to quantify through observing his behaviors directly\cite{hu2008collaborative}. More than this, there are even some unexpected behaviors in the historical interactions. For examples, as shown in figure \ref{fig: unexpected_behaviors}, a businessman has bought a baby T-shirt for his son, and clicked a handicraft by accident or due to curiosity, as shown in red boxes. Consider another scenario, someone has watched a bunch of movies, but not every movie was what he likes. In these two cases, we argue that these unexpected behaviors are irrelevant to the user's preference. It will hurt the recommendation performance dramatically when we can not exploit these behaviors in a discriminating way. That means we need to eliminate the effects of unexpected behaviors. However, existing works do not have clear information to guide the model to eliminate the effects of unexpected behaviors and they achieve inferior performances.

In this work, we aim to fill the research gap by developing a solution that eliminates the effects of unexpected behaviors under the guidance of the target item. Towards this end, we propose a novel method, named \textbf{M}ulti \textbf{P}reference \textbf{M}odel (MPM), which can not only eliminate the effects of unexpected behaviors but also retain the predictive information as much as possible. Specifically, we first extract the users' instant preferences from their recent historical interactions by a fine-grained preferences module. And then we train an unexpected-behaviors detector to judge whether these instant preferences are biased by unexpected behaviors. This is done under the guidance of the target item, as the instant preferences should stay in line with the positive target items and keep a distance from the negative ones. We also integrate a general preference module, which is a neural latent factor model\cite{he2017neural} based on collaborative filtering. The integration of the general preference module in our model can ensure the recommendation will not deviate far away from the users' real general preferences. Thereafter, an output module is  performed to eliminate the effects of unexpected behaviors and aggregate all information to obtain the final recommendation.  We conduct extensive experiments on two datasets about movie and e-retailing to verify our method, demonstrating significant improvements in our Multi-Preferences Model over the state-of-the-art methods. MPM gets a massive improvement in HR@10 and NDCG@10, which relatively increased by 3.643\% and 4.107\% compare with AttRec model on average.

The contributions of this work are threefold:

\begin{enumerate}
	\item We highlight the importantance of eliminating the unexpected behaviors' effect in historical interactions, which is crucial for recommendations.
	\item We propose an end-to-end neural network model which can eliminate the effects of unexpected behaviors in historical interactions.
	\item We conduct experiments on public datasets to demonstrate the effectiveness of the proposed model, and it outperform the competitive baselines.
\end{enumerate}


\section{Related Work}

As a personalized filtering tool, recommender systems have been widely adopted to alleviate the information overload problem. In this section, we briefly review related works from three perspectives: session-based recommendation, latent factor models and a new trend in recommendation.

\subsection{Session-based Recommendation}

In the session-based setting, recommender systems have to rely only on the interactions of the user in the current sessions to provide accurate recommendations\cite{hidasi2018recurrent}. And, our work mainly integrates the recent historical interactions to the conventional latent factor model, meanwhile we eliminate the effects of unexpected behaviors.

In the session-based recommendation, Hidasi \textit{et al.}\cite{hidasi2016session-based} take the lead in exploring GRUs for the prediction of the next user's action in a session. Li \textit{et al.}\cite{li2017neural} argue that both the user's sequential behaviors and the main purpose in the current session should be considered in recommendations. Tuan \textit{et al.}\cite{tuan20173d} describe a method that combines session clicks and content features to generate recommendations. Ruocco \textit{et al.}\cite{ruocco2017inter-session} use a second RNN to learn from recent sessions, and predict the user's interest in the current session. Wu \textit{et al.}\cite{wu2017session-aware} incorporate different kinds of user search behaviors such as clicks and views to learn the session representation, which can get a good result in recommendations. Session-based recommendation systems share many similarities with our work, they both can take the advantage of historical interactions for recommendations.

\subsection{Latent Factor Models}

Latent factor models get its name from mapping users and items to latent factor vectors. In the age of machine learning, latent factor models are first proposed based on matrix factorizations to deliver accuracy superior to classical nearest-neighbor techniques\cite{koren2009matrix}. Latent factor models are canonical for capturing  user's general preference, as they are trained by considering the whole historical interactions. After entering the era of deep learning, latent factor model has been developed in different ways\cite{batmaz2018review}.

Salakhutdinov \textit{et al.}\cite{salakhutdinov2007restricted} first use Restricted Boltzmann Machine to extract latent features of user preferences from the user-item preference matrix. In\cite{sedhain2015autorec}, Autoencoder is used in deep latent factor models to learn a non-linear representation of the user-item matrix. He \textit{et al.}\cite{he2017neural} replace the inner product with an MLP as the interaction function. They present a general framework named NCF and implement a GMF and NeuMF model under the NCF framework. Travis \textit{et al.}\cite{ebesu2018collaborative} integrate the latent factor model and the neighborhood-based model by memory network and propose the collaborative memory network model. In our work, we integrate a latent factor model to capture the user's general preference, which endows our model with particular robustness.

\subsection{New Trend in Recommendation}

From the literature, we can find that session-based recommender systems produce recommendations base on short-term preference and latent factor models mainly rely on long-term (general) preference. Recently, a surge of interest in combining long-term and short-term preference for recommendation has emerged.  Liu \textit{et al.}\cite{liu2018stamp:} propose a novel short-term attention/memory priority model (STAMP), which is capable of capturing users' general interests from the long-term memory of a session context, whilst taking into account users' current interests from the short-term memory ofthe last-clicks. BINN learns historical preference and present motivation of the target users using two LSTM-based architecture by discriminatively exploiting user behaviors\cite{li2018learning}. Zhang \textit{et al.}\cite{zhang2018next} propose an AttRec model which learns the short-term intention by self-attention module and extract long-term intention in the traditional manner which measure the closeness between item $i$ and user $u$ by Euclidean distance. These works have achieved considerable success. However, as mentioned before, there are usually some unexpected behaviors in historical interactions, and these works do not have clear information to guide the model to learn to eliminate the effects of unexpected behaviors. Towards this end, we propose a Multi-Preferences Model to address this problem.


\section{Multi-Preferences Model}

\begin{figure*}[ht]
	\vspace{-0.4cm}
	\setlength{\abovecaptionskip}{0.2cm}
	\setlength{\belowcaptionskip}{-10pt}
	\includegraphics[width=\linewidth]{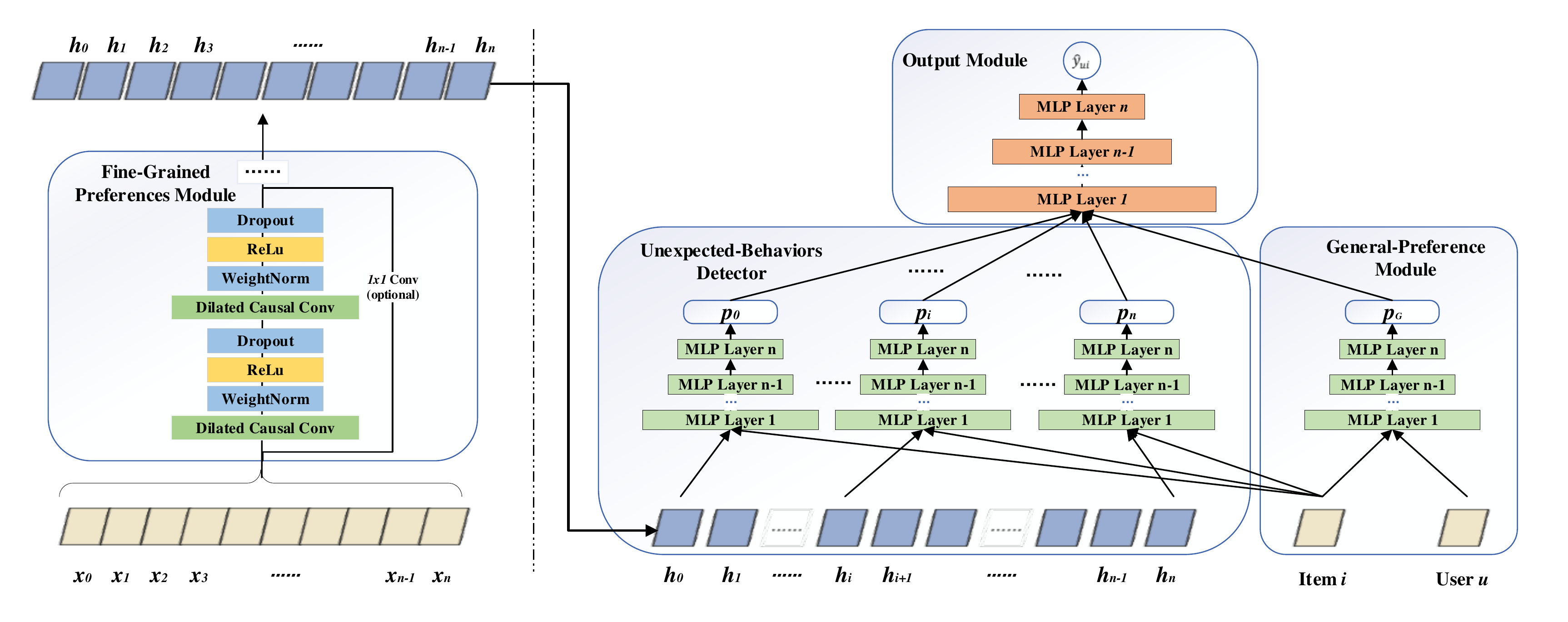}
	\caption{The framework of Multi-Preferences Model. (1) Fine-grained preferences module learns the instant preferences from historical interactions, (2) unexpected-behaviors detector detect the unexpected behaviors, (3) general-preference module capture user's general preference, (4) output module eliminate the effects of unexpected behaviors and produce final recommendation.}
	\label{fig: framework}
\end{figure*}

In this section, we will elaborate on the multi-preference model, as illustrated in figure \ref{fig: framework}. Here, we first formally define the recommendation problem in implicit feedback datasets and then present MPM in detail.

\subsection{Recommendation in Implicit Feedback Datasets}

In the implicit feedback setting, the interactions between users and items such as purchase, browse, click, view or even mouse movements are informative for recommendations. However, users' opinion can't be deduced directly through observing these behaviors\cite{hu2008collaborative}. This brings great challenges to the recommendation problem.  

We assume there are $M$ users and $N$ items in the dataset and define the user-item interaction matrix $Y=R^{M * N}$ as equation \ref{eq. implicit_feedback}. If exist interactions between user \textit{u} and item \textit{i} set $y_{ui}=1$, otherwise, $y_{ui}=0$ .

\begin{equation}
\centering
\setlength{\abovedisplayskip}{2pt}
\setlength{\belowdisplayskip}{2pt}
y_{ui}=
\begin{cases}
1\ \text{if interactions (user \textit{u}, item \textit{i}) are observed;}\\
0\  \text{otherwise.}
\end{cases}
\label{eq. implicit_feedback}
\end{equation}

Our task can be formulated as follows: Given a user $u$, a target item $i$, and the last $K$ interactions $H^{ui}=[x_{0},x_{1},x_{2},...,x_{K}]$. The holistic goal is to estimate the interaction by:

\begin{equation}
\centering
\hat{y}_{ui} = f_{\Theta}(u,i|H^{ui})
\label{eq. abs_function}
\end{equation}

Where $f$ denotes the underlying model with parameters $\Theta$, and $\hat{y}_{ui}$ presents the predicted score for the interaction. We can explain $\hat{y}_{ui}$ as the likelihood of the user $u$ interact with the item $i$ based on the recent historical interactions $H^{ui}$.

\subsection{Modeling}

Multi-preferences model takes the user's last $K$ historical interactions as input and outputs a score indicating how possible the user will interact with the target item. As illustrated in figure \ref{fig: framework}, there are four key components in MPM: (1) fine-grained preferences module learn the fine-grained instant preferences from historical interactions, (2)  unexpected-behaviors detector validate whether the instant preferences are biased by unexpected behaviors, (3) general-preference module is a conventional neural latent factor model for capturing general preference,(4) at last, the output module is performed to eliminate the effects of unexpected behaviors and fuse all information to produce a final recommendation.

\subsubsection{Fine-Grained Preferences Module}

Given the recent historical interactions, we need to extract useful knowledge for recommendations. Instead of learning a single representation of the historical interactions in an autoencoding manner like previous works\cite{zhang2018next}, we output an instant preference in each step. This workflow is consistent with the conventional sequential modeling, which obey the autoregressive principle. Here, we implement the fine-grained preference module based on the Temporal Convolutional Network.

Temporal Convolutional Network (TCN) is a popular architecture. Studies have shown that TCN outperforms canonical recurrent networks such as LSTMs across a diverse range of tasks and datasets\cite{bai2018an}. More formally, an input of $1-D$ items sequence $H^{ui} \in R^{n}$ and a filter $f:\left\{0,...,k-1\right\} \rightarrow R$, the dilated convolution operation $F$ on element $x_{s}$ of the sequence is defined as

\begin{equation}
F(x_{s}) = (H^{ui} *_{d} f)(x_{s}) = \sum_{i=0}^{k-1}f(i).H^{ui}_{s-di}
\label{eq. dilated_convolution}
\end{equation}

\begin{figure}[h!]
	\centering
	\vspace{-0.4cm}
	\setlength{\abovecaptionskip}{0.2cm}
	\setlength{\belowcaptionskip}{-10pt}
	\includegraphics[width=0.7\linewidth]{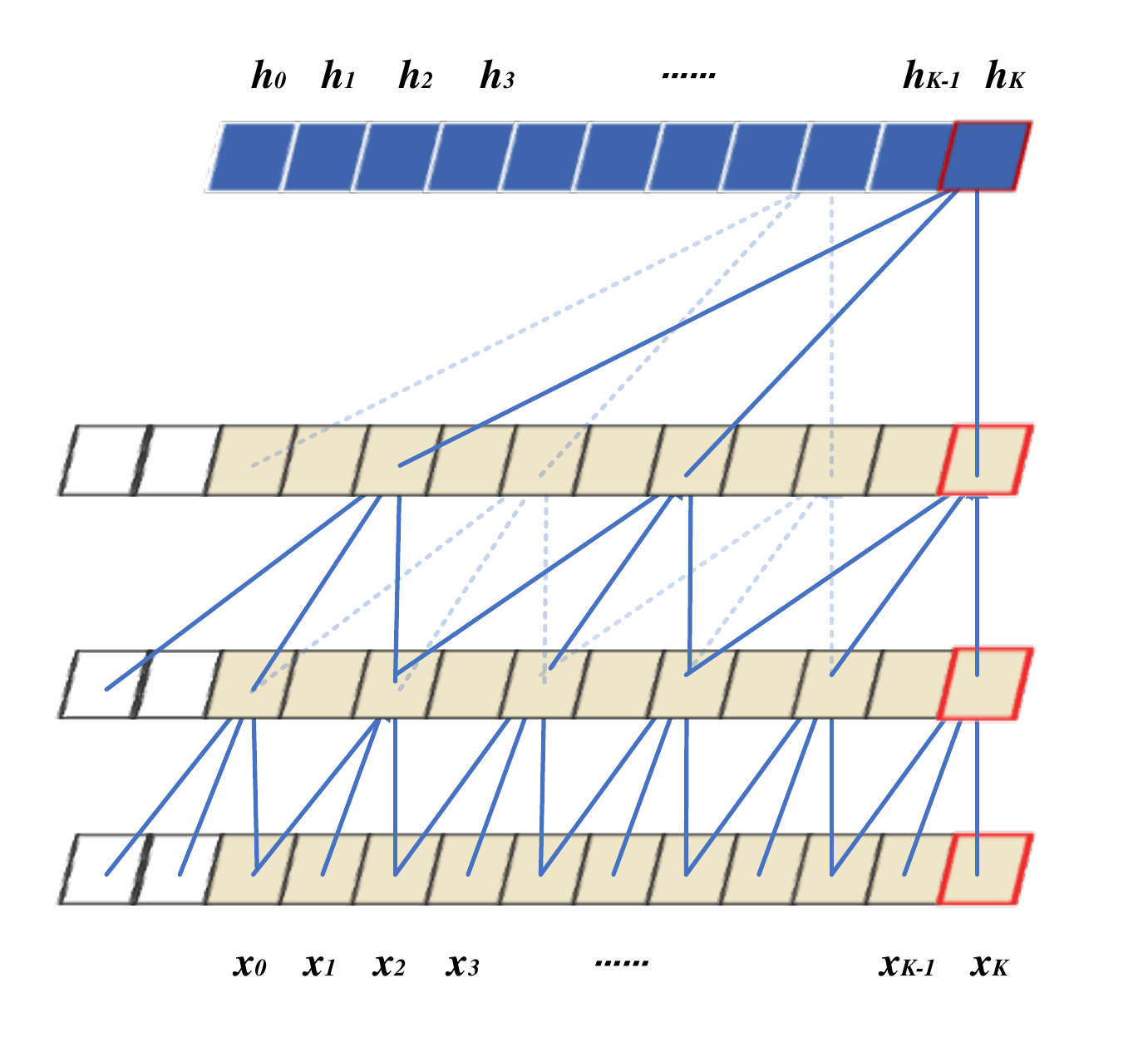}
	\caption{A dilated convolution with dilation factors $d = 1,2,4$ and filter size $k = 3$.}
	\label{fig: dilated_causal_convolution}
\end{figure}

Where $d$ is the dilation factor, $k$ is the filter size, and $s-di$ accounts for the direction of the past. Dilated convolution is illustrated in figure \ref{fig: dilated_causal_convolution}. And zero padding of length $(k - 1)$ is added to keep subsequent layers as the same length as previous ones. TCN also incorporates residual connection\cite{he2016deep}, weight normalization\cite{salimans2016weight} and spatial dropout\cite{srivastava2014dropout} to improve performance, as shown in figure \ref{fig: framework} (a). In our model, the fine-grained preferences module can be abstracted as equation \ref{eq: TCN}

\begin{equation}
[h_{0},...,h_{i},...,h_{K}] = TCN(x_{0},x_{1},x_{2},...,x_{K})
\label{eq: TCN}
\end{equation}

Where $h_{i}$ contains all the semantic information of the interactions before position $i$, that is $h_{i} = \Psi(x_{0},x_{1},...,x_{i})$. $h_{i}$ represents the user's instant preference at this moment. However, some of these instant preferences always are biased as the existence of unexpected behaviors in historical interactions. For example, $x_{i}$ is a gift that the user buys for his friend or he click the item by accident. As a consequence, some instant preferences will be destroyed by unexpected behaviors. 

\subsubsection{Unexpected-Behaviors Detector}

To eliminate the effects of unexpected behaviors, we firstly builds an unexpected-behaviors detector to judge whether an instant preference is biased by unexpected behaviors. More specifically, in the training phase, unexpected-behaviors detector incorporates the target items to judge whether an instant preference is biased, as the instant preferences should stay in line with the positive target items and keep a distance from the negative ones. Once the model trained, the detector has the ability to make a judgment by itself. 

\begin{figure}[h!]
	\centering
	\vspace{-0.4cm}
	\setlength{\abovecaptionskip}{0.2cm}
	\setlength{\belowcaptionskip}{-10pt}
	\includegraphics[width=\linewidth]{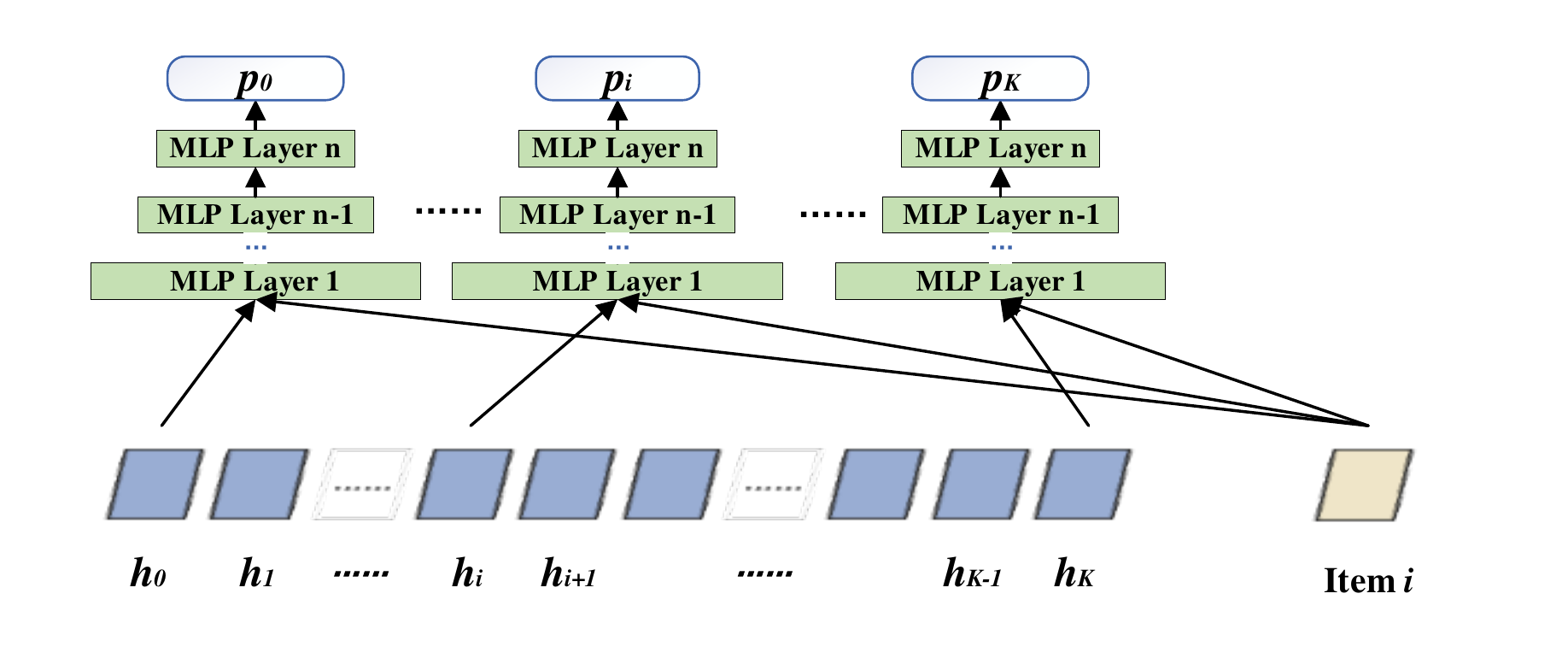}
	\caption{Unexpected-Behaviors Detector.}
	\label{fig: unexpected_behaviors_detector}
\end{figure}

The unexpected-behaviors detector is shown in figure \ref{fig: unexpected_behaviors_detector}, which consist of $K$ multi-layers perceptron (MLP).

Every instant preference $h_{i}$ will be concatenated with a target item embedding $e_{i}$, and then send into an MLP. The MLP can model the interactions between instant preference and item latent features. Finally, it outputs a predictive vector $p_{i}$, which contains the information about whether the instant preference is distorted. Unexpected-behaviors detector is defined as follows.

\begin{align}
\setlength{\abovedisplayskip}{-3pt}
\setlength{\belowdisplayskip}{3pt}
p_{i} = z_{n} &=  a_{n}(W{_n-1^T}z_{n-1} + b_{n-1})\\
z_{n-1} &=  a_{n-1}(W_{n-2}^{T} z_{n-2} + b_{1})\\
... \notag \\
z_{2} &=  a_{2}(W{_1^T}z_{1} + b_{1})\\
z_{1} &=  \left[
\begin{matrix}
h_{i}\\
e_{i}
\end{matrix}
\right]
\label{eq. MLP}
\end{align}

Where $W_{x}$, $b_{x}$, and $a_{x}$ represent the corresponding weight matrix, bias, and activation function in the $x-th$ layer of the MLP respectively. All in all, unexpected-behaviors detector consumes the instant preferences and outputs a bunch of predictive vectors $P=[p_{0},p_{1},...,p_{K}]$, which indicate whether the instant preference is distorted.

\subsubsection{General Preference Module}

So far, we have described the unexpected-behaviors detector in detail and the predictive vectors $P$ are informative for recommendations. However, due to the complexity of the real world, we still need to introduce general preference to make sure recommendations go in the right direction. Here, we build a general preference module base on conventional neural latent factor model\cite{he2017neural}, which is proficient in capturing the general preference by considering the whole user-item matrix.

\begin{figure}[h!]
	\centering
	\vspace{-0.4cm}
	\setlength{\abovecaptionskip}{0.2cm}
	\setlength{\belowcaptionskip}{-10pt}
	\includegraphics[width=0.8\linewidth]{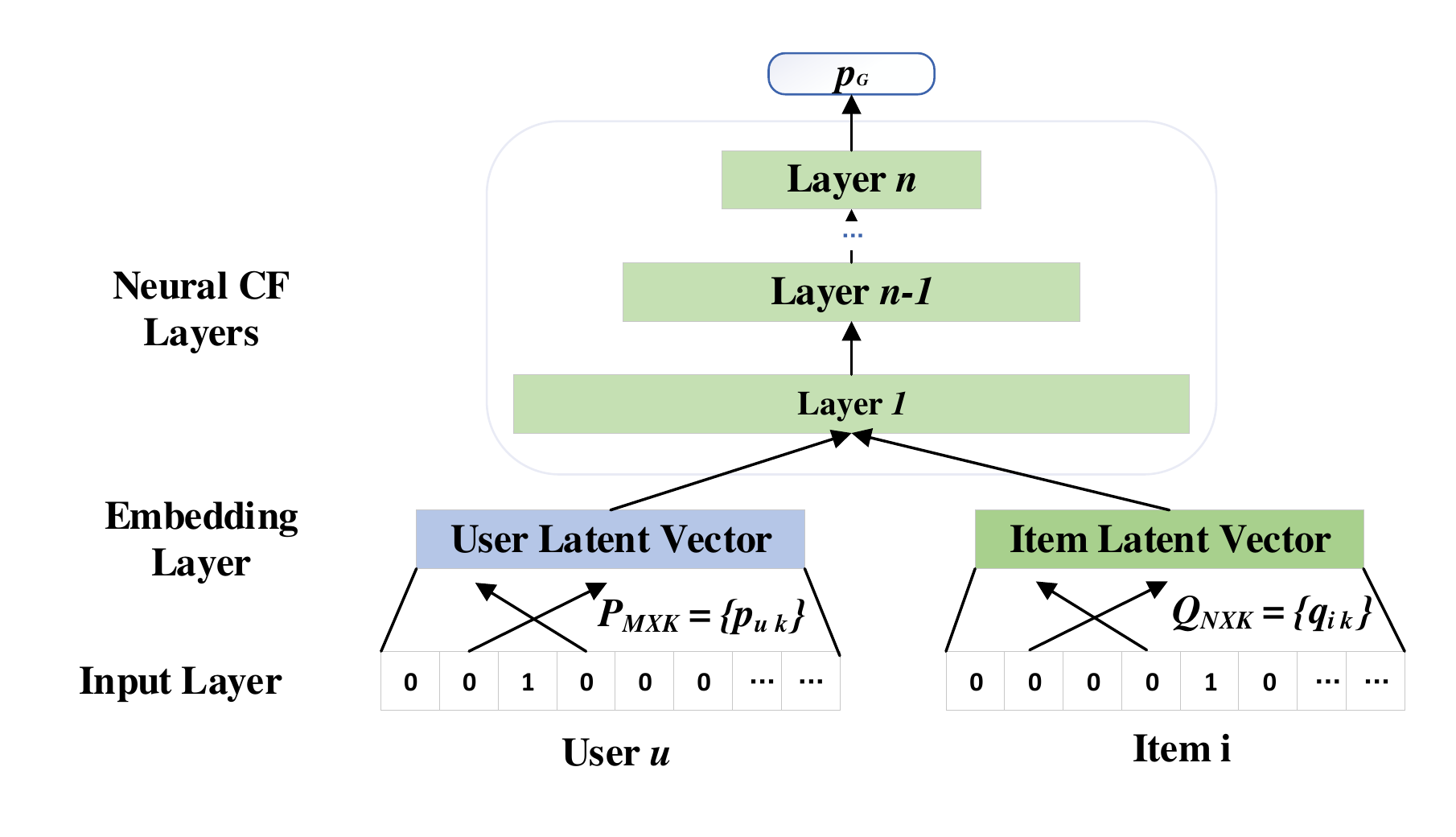}
	\caption{General Preference Module.}
	\label{fig: General_Preference_Module}
\end{figure}

General preference module takes the user latent vector $e_{u}$ and item latent vector $e_{i}$ as input and outputs a predictive vector $p_{G}$, as illustrated in figure \ref{fig: General_Preference_Module}. As the user latent factor vector contains the user's general preference and item latent factor vector contains the characteristics of the item $i$\cite{koren2009matrix}, the predictive vector $p_{G}$ indicates whether the target item $i$ fit the user's general preference. 

The integration of the General Preference Module in our model can ensure the recommendation will not deviate from the general preferences of users. It can improve the robustness of the model by a large margin. The general preference module is defined as follows,

\begin{equation}
p_{G} = a_{n}(W_{n}(a_{n-1}(...a_{1}(W_{1}\left[e_{u},e_{i}\right] + b_{1})...)) + b_{n})
\label{eq. General_MLP}
\end{equation}

\subsubsection{Output Module}

\begin{figure}[h!]
	\centering
	\vspace{-0.4cm}
	\setlength{\abovecaptionskip}{0.2cm}
	\setlength{\belowcaptionskip}{-10pt}
	\includegraphics[width=\linewidth]{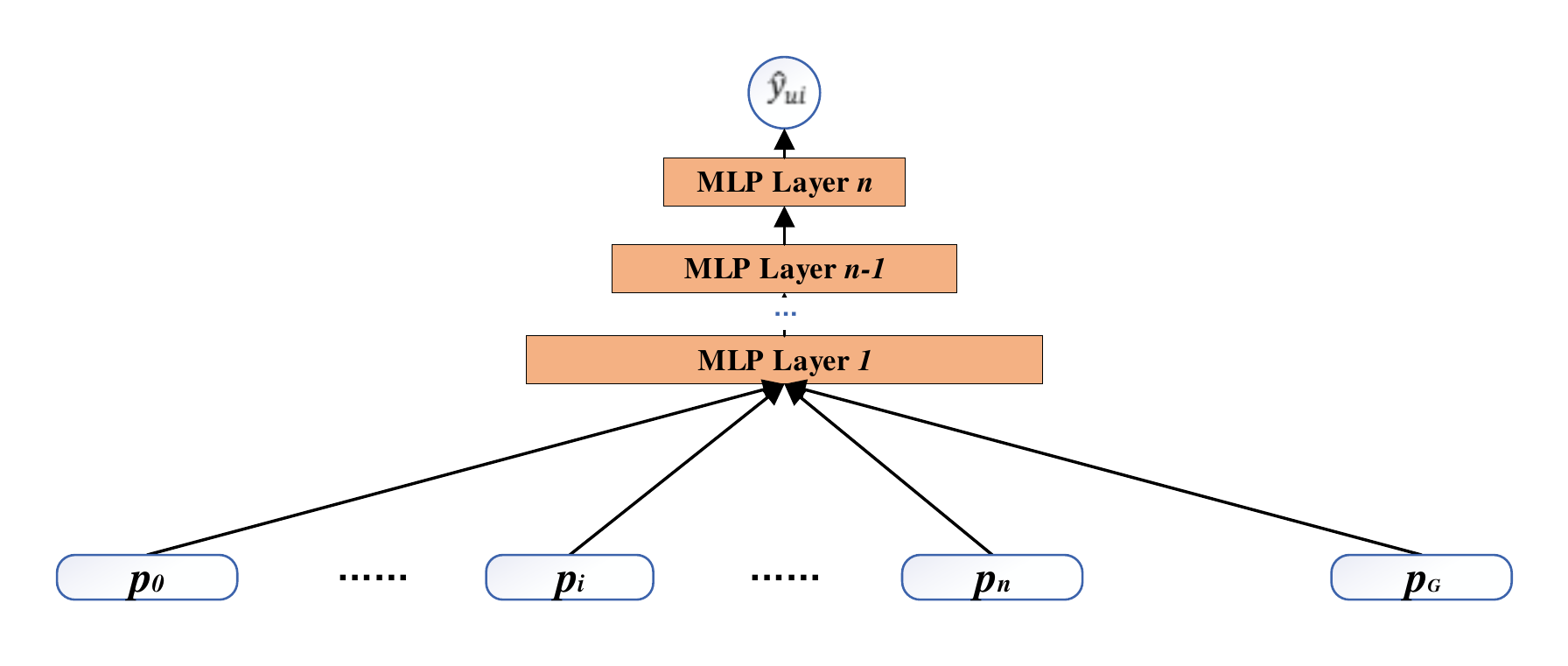}
	\caption{Output Module. It eliminates the effects of unexpected behaviors and outputs the final prediction.}
	\label{fig: Output_Module}
\end{figure}

Until now, we still haven't eliminated the effects of unexpected behaviors yet. In our model, the Output Module has two important tasks: (1) eliminates the effects of unexpected behaviors base on the detection of unexpected-behaviors detector as much as possible, (2) integrates all the information to make a final prediction. We build our Output Module atop MLP, it will automatically learn how to complete these two mentioned tasks. MLP is a powerful architecture, it is sufficient to represent any function in theory\cite{Goodfellow-et-al-2016}. The Output Module is illustrated in figure \ref{fig: Output_Module}.

Output module consumes the predictive vectors $P=[ p_{0}, p_{1}, p_{2}, ..., p_{K}]$ to eliminate the effects of unexpected behaviors and integrate $p_{G}$ make the final prediction $\hat{y}_{ui}$, as defined in equation \ref{eq. Output_MLP}. $\hat{y}_{ui}$ indicate the likelihood of user $u$ interacts with the item $i$ in the next time.

\begin{equation}
\hat{y}_{ui} = a_{n}(W_{n}(a_{n-1}(...a_{1}(W_{1}\left[p_{0},p_{1},...,p_{K},p_{G}\right] + b_{1})...)) + b_{n})
\label{eq. Output_MLP}
\end{equation}

\subsection{Learning}

Similar to the spirit in recent works\cite{he2017neural}, the recommendation problem in implicit feedback datasets can be regarded as a binary classification problem, where an observed user-item interaction is assigned a target value 1, otherwise 0. We choose the binary cross-entropy as our loss function, and learn the parameters of MPM in a pointwise learning manner. The loss function is defined as equation \ref{eq. loss}.

\begin{equation}
loss = - \sum_{(u,i) \in y \cup y^{-}}{ylog\hat{y}_{ui} + (1 - y)log(1 - \hat{y}_{ui})}
\label{eq. loss}
\end{equation} 

Here $y$ denotes the positive sample set and $y^{-}$ means the negative sample set. We elaborate on the implementation details in the section of Experimental Settings.


\section{Experiments}

In this section, we conduct experiments on two real-world datasets to evaluate our proposed method. We aim to answer the following research questions:

\begin{enumerate}
	\item \textbf{RQ1:} Compared with the state-of-the-art methods in implicit feedback setting, how does our method perform?
	\item \textbf{RQ2:} How do recent historical interactions affect the performance of our method?
	\item \textbf{RQ3:} Will a larger model size be helpful for improving performance?
\end{enumerate}

\subsection{Dataset Description}

We consider two scenarios: movie recommendation and e-retailing recommendation. 

\begin{enumerate}
	\item \textbf{For movie domain:} we use the MovieLens Datasets\cite{harper2016movielens}, which are typically used as a benchmark dataset for the recommender systems. We chose to experiment with the ml-20m, ml-10m and ml-1m subsets of MovieLens.
	\item \textbf{For e-retailing domain:} we use the Taobao User-Behavior Dataset\cite{zhu2018learning}, which is a behavior dataset provided by Alibaba for the study of implicit feedback recommendation problem. We sample three subsets from Taobao user-behavior data set, denoted as Taobao-1m, Taobao-10m, and Taobao-20m.
\end{enumerate}

Detail statistics of the datasets are presented in Table \ref{tab: datasets}.

\begin{table}[h!]
	\caption{Statistics of the datasets used in experiments}
	\begin{center}
		\begin{tabular}{c c c c}
			\toprule
			\textbf{Dataset} &	\textbf{Users}	& \textbf{Items} & \textbf{Interactions}
			\cr \midrule
			ml-1m &	6040 &	3706 &	1000209\\
			ml-10m &	69878 &	10677 &	10000054\\
			ml-20m &	138493 & 26744 & 20000263\\
			Taobao-1m &	9879 &	402255 & 993980\\
			Taobao-10m & 98799&	1590861  &	10004137\\
			Taobao-20m &	197598 & 2215070 &	20039836\\
			\bottomrule
		\end{tabular}
		\begin{tablenotes}
		\centering
		\item Note: filter out users with less than 20 interactions.
		\end{tablenotes}
		\label{tab: datasets}
	\end{center}
\end{table}

Following previous efforts\cite{he2016fast,ebesu2018collaborative,he2017neural}, we evaluate the performance of our proposed model by the leave-one-out evaluation. For each dataset, we hold out the last one item that each user has interacted with and sample 99 items that unobserved interactions to form the test set, a validation set is also created like the test set and remaining data as a training set. For each positive user-item interaction pair in the training set, we conducted the negative sampling strategy to pair it with four negative items. Our datasets are formed as follows,

\begin{equation}
\left\{ Uer\ u, Item\ i, H^{ui} = (x_{0},x_{1},x_{2},...,x_{n}), y_{ui}\right\} 
\end{equation}
\begin{equation}
y_{ui} = \left\{0, 1\right\}
\end{equation}

\subsection{Experimental Settings}

\subsubsection{Evaluation Metrics}

We rank the test item with the 99 negative items and then select the $top-K$ items as a recommendation list. We evaluate our model by  HR (Hit-Rate) and NDCG (Normalized Discounted Cumulative Gain) metrics, which are defined as follows: 

\begin{align}
HR@K &= \dfrac{1}{M}\sum_{u \in U} 1 (R_{u,i} \leq K)\\
NDCG@K &= \dfrac{1}{M}\sum_{u \in U} \dfrac{1}{IDCG_{p}} \sum_{i=1}^{p} \dfrac{2^{rel_{i}}-1}{log_{2} (i+1)}\\
IDCG_{p} &= \sum_{i=1}^{|REL_{p}|} \dfrac{2^{rel_{i}}-1}{log_{2} (i+1)}	
\end{align}

Here, In the HR metric, $R_{u,i}$ is the rank generated by the model for item $i$. If a model ranks $i$ among the $top-K$, the indicator function will return 1, otherwise 0. In the NDCG, $p$ is the rank position of $i$ in the recommendation list, and IDCG means the ideal discounted cumulative gain\cite{wang2013theoretical}. Intuitively, HR measures the presence of the positive item within the $top-K$ and NDCG measures the item's position in the ranked list and penalizes the score for ranking the item lower in the list.

\subsubsection{Baselines}

We compare our proposed approach against the competitive baselines list below.

\begin{itemize}
	\item \textbf{MF}\cite{Rendle2011Factorization}, a general matrix factorization model which is one of the most effective techniques for capturing long-term preferences.
	\item \textbf{NeuMF}\cite{he2017neural} is a composite matrix factorization jointly coupled with a multi-layer perceptron model for item ranking.
	\item \textbf{AttRec}\cite{zhang2018next} take both short-term and long-term intentions into consideration. We use an MLP network replace Euclidean distance to measure the closeness between item $i$ and user $u$. 
	\item\textbf{MPM-attn}, we also implemented a simple version of the MPM model in the conventional fashion\cite{zhang2018next}, in which we learn a short-term preference from instant preferences by self-attention mechanism\cite{lin2017a}, and then combine the short-term preference and general preference for recommendations. 
\end{itemize}

\subsubsection{Parameter Settings}

For a fair comparison, we learn all models by binary cross-entropy loss and optimize all models with Adaptive Moment Estimation--Adam\cite{kingma2015adam:}. We apply a grid search to find out the best settings of hyperparameters. After a grid search was performed on the validation set, the hyperparameters were set as follows, the $learning-rate$ was set to 0.001, $beta_{1}$ was set to 0.9, $beta_{2}$ was set to 0.999, $epsilon$ was set to 1e-8, $embedding-size$ was set to 32 and $batch-size$ was set to 1024. We employed a four layers architecture $[64,128,64,32]$ for multi-layer perceptron. In the TCN module, the number of dilated convolution layers is set to 4, our kernel size is 3 and the dilation factors is $[1,2,4,8]$.

\subsection{Performance Comparison (RQ1)}

Table \ref{tab: performance} reports our experimental results for HR and NDCG with cut off at 10 on the MovieLens and Taobao User-Behavior datasets, in which the number of historical interactions (history size) is set to 9.

\begin{table*}[htb]
	
	\caption{Experimental results for different methods on the MovieLens and Taobao datasets.}
	\begin{center}
		\begin{threeparttable}
			\begin{tabular}{c c c c c c c c}
				\toprule
				\textbf{Dataset}              &\textbf{Metric}  &\textbf{MF}  &\textbf{NeuMF}   &\textbf{AttRec}     &\textbf{MPM-attn} &\textbf{MPM}	  &\textbf{Improv}
				\cr \midrule
				\multirow{2}{*}{ml-20m}        &HR@10  	 &0.6120	    &0.6330 		  &0.7321 		         &0.7289 	     &\textbf{0.7555}      &\textbf{3.20\%} \\
				&NDCG@10  	&0.3644	&0.3798           &0.4823                &0.4793         &\textbf{0.5128}      &\textbf{6.13\%} \\
				\multirow{2}{*}{ml-10m}        &HR@10      &0.8798      &0.8833           &0.9223                &0.9229         &\textbf{0.9319}      &\textbf{1.04\%} \\
				&NDCG@10    &0.6227      &0.6298           &0.7140                &0.7132         &\textbf{0.7385}      &\textbf{3.43\%} \\
				\multirow{2}{*}{ml-1m}         &HR@10    &0.7088        &0.7018           &0.8069                &0.8153         &\textbf{0.8218}      &\textbf{1.85\%} \\
				&NDCG@10     &0.4296     &0.4247           &0.5788                &0.5962         &\textbf{0.6065}      &\textbf{4.79\%} \\
				\multirow{2}{*}{Taobao-20m}    &HR@10    &0.7122        &0.7461           &0.7507                &0.7641         &\textbf{0.7918}      &\textbf{5.47\%} \\
				&NDCG@10     &0.5156     &0.5923           &0.6034                &0.5958         &\textbf{0.6634}      &\textbf{3.25\%} \\
				\multirow{2}{*}{Taobao-10m}    &HR@10      &0.6668      &0.6878           &0.6981                &0.6884         &\textbf{0.7218}      &\textbf{9.34\%} \\
				&NDCG@10     &0.5257     &0.5661           &0.5579                &0.5248         &\textbf{0.5972}      &\textbf{7.04\%} \\
				\multirow{2}{*}{Taobao-1m}     &HR@10     &0.4525       &0.4686           &0.4812                &0.4779         &\textbf{0.4858}      &\textbf{0.96\%} \\
				&NDCG@10     &0.3865     &\textbf{0.4014}           &0.3964                &0.3791         &0.3918      &\textbf{-\%} \\
				
				\bottomrule
			\end{tabular}
			\begin{tablenotes}
				\item Note: \textit{The best results are highlighted in \textbf{bold}. The \textbf{Improv} was computed compare with the AttRec model.}
			\end{tablenotes}
		\end{threeparttable}
		\label{tab: performance}
	\end{center}	
\end{table*}

From the experimental results, we have the following findings:

\begin{itemize}
	\item \textbf{MF} gives poor performance in both datasets. This indicates that the conventional latent factor models, which rely heavily on the general preference, may fail to capture the slight variation in the preference and have not considered the evolution of users' preferences. As a consequence, they achieve an inferior performances.
	\item \textbf{NeuMF} outperform MF, which demonstrates that incorporating an MLP to model the interaction between users and items is beneficial. NeuMF learn an arbitrary function from data by replacing the inner product with a neural architecture.
	\item \textbf{AttRec} achieves better performance than NeuMF. It makes sense since by incorporating short-term intention learned from recent historical interactions as fine adjustment. AttRec fully explore both long-term intent and short-term preference for recommendations. 
	\item It is worth mentioning that \textbf{MPM-attn} also achieves comparable performance to AttRec. MPM-attn is an alternative choice for integrating both long-term preference and short-term preference to make a recommendation, it extracts user's short-term preference from the recent historical interactions by an self-attention mechanism\cite{lin2017a}.
	\item \textbf{MPM} substantially outperforms AttRec $w.r.t$ HR@10 and NDCG@10, achieving the best performance. By  introducing the positive interactions which are in conformity with the user's real preferences to train the model, after training, MPM has the ability to eliminate the effect of unexpected behaviors. MPM gets a massive improvement in HR@10 and NDCG@10, which relatively increased by 3.643\% and 4.107\% on average. This demonstrate the importance of eliminating the effects of unexpected behaviors. It also illustrates the effectiveness of proposed model.
\end{itemize}

\subsection{Study of MPM (RQ2)}

For a comprehension of our method, we dive into an in-depth model analysis. We start by exploring the influence of multi-preferences to investigate whether MPM can eliminate the effects of unexpected behaviors. We then study how the history size affects the performance. We only conduct experiments on the Taobao-20m dataset for saving space.

\subsubsection{Impact of Multi-Preferences}

To our knowledge, there are two methods to integrate the short-term preference and long-term preference for recommendations in the literature. But, actually, the main difference between them is the way of learning short-term preference. One learns a single representation of short-term preference from the recent historical interaction in an autoencoding manner directly\cite{zhang2018next}, in which it doesn't need to follow the principle of autoregression. The other extracts instant preferences by sequential model in an autoregression manner and then encode these instant preferences to obtain the short-term preference by self-attention mechanism, just like MPM-attn does.

\begin{figure}[h!]
	\centering
	\setlength{\abovecaptionskip}{0.2cm}
	\setlength{\belowcaptionskip}{-10pt}
	\includegraphics[width=0.8\linewidth]{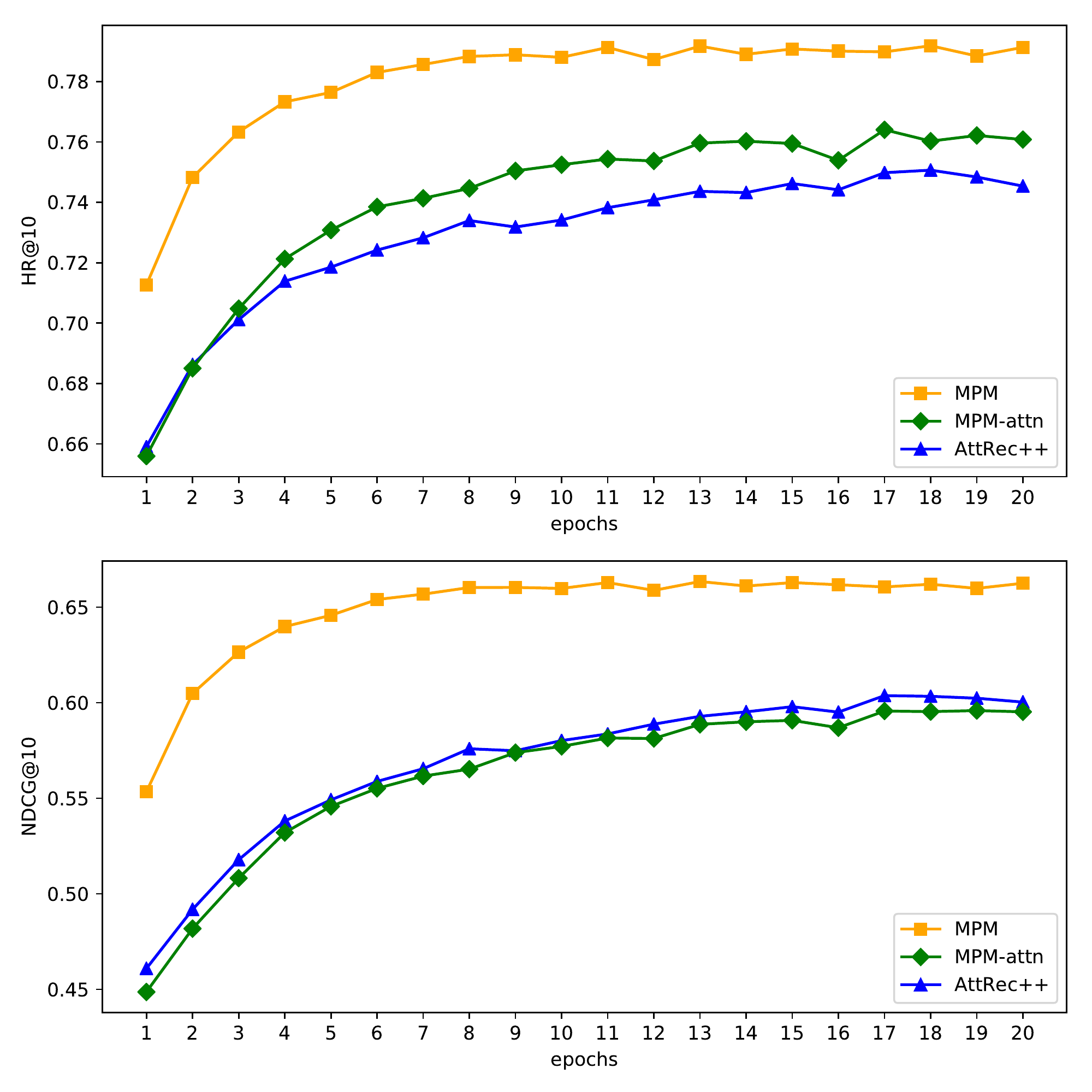}
	\caption{Impact of Multi-Preferences}
	\label{fig. effects_of_multi_preferences}
\end{figure}

However, these works fail to eliminate the effect of unexpected behaviors, as no information to guide model learning. To address this problem, we propose a Multi-Preferences Model, it eliminates the effect of unexpected behaviors through a multi-preferences way. We explore the influence of multi-preferences on Taobao-20m dataset, as illustrated in figure \ref{fig. effects_of_multi_preferences}.

From figure \ref{fig. effects_of_multi_preferences}, we can see that MPM-attn achieved comparable performance to AttRec, even outperforms slightly. After eliminating the effects of unexpected behaviors, the Multi-Preferences Model substantially outperforms the state-of-the-art method by a larger margin in both HR@10 and NDCG@10, which increased by 5.47\% and 3.23\%. This demonstrates that unexpected behaviors can hurt the recommendation performance dramatically and our model can eliminate the effects of unexpected behaviors. 

\subsubsection{Impact of History Size}

\begin{figure}[h!]
	\centering
	\setlength{\abovecaptionskip}{0.2cm}
	\setlength{\belowcaptionskip}{-10pt}
	\includegraphics[width=0.8\linewidth]{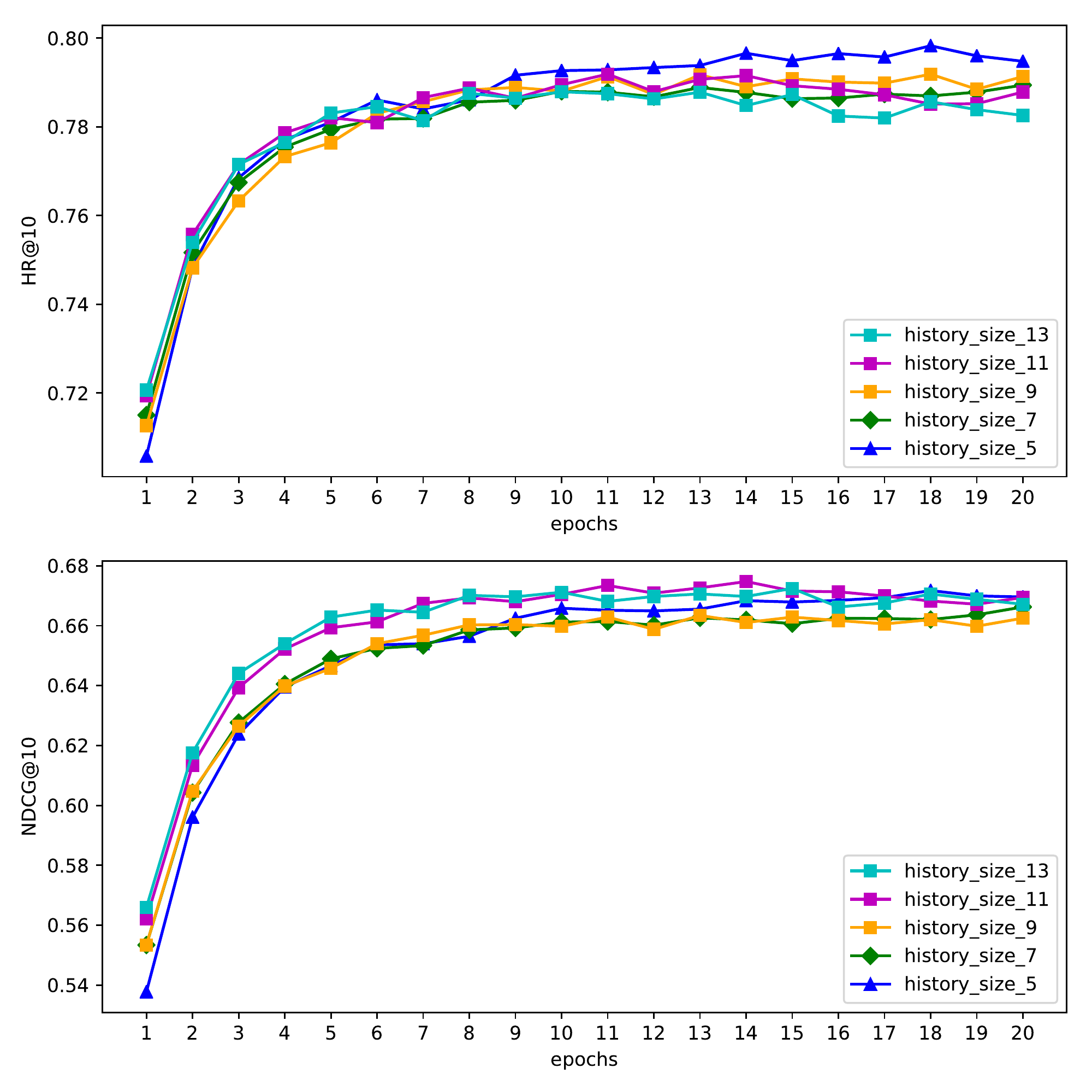}
	\caption{Impact of History Size}
	\label{fig. effects_of_history_size}
\end{figure}

The other important thing about our model is to decide how many historical interactions should be considered. We cannot simply assume that the longer the history,  the more information MPM can capture,  and the better performance we got. As users' preferences evolve over time, too big a history size can only add the burden on the model and bring in some noise. Here, we study the effect of history size on Taobao-20m dataset, we search the history size on $\left\{5,7,9,11,13\right\}$, as illustrated in figure \ref{fig. effects_of_history_size}.

From figure \ref{fig. effects_of_history_size}, we can't observe significant differences between the performances as history size growing from 5 to 13. However, looking from the other side, this can be saw as evidence of MPM can eliminate the effects of unexpected behaviors efficiently, as the larger of history size, the more unexpected interactions it will encounter possibly, but the performance still keeps in a reasonable region. In addition, we can see a slight improvement on NDCG@10 metric. This indicates that we get a higher quality recommendation list from the model. But, in practices, we should tune the history size for different dataset, this is a trade-off between performance and complexity.

\subsection{Is Larger Model Size Helpful ? (RQ3)}

\begin{figure}[h!]
	\centering
	\setlength{\abovecaptionskip}{0.2cm}
	\setlength{\belowcaptionskip}{-10pt}
	\includegraphics[width=0.8\linewidth]{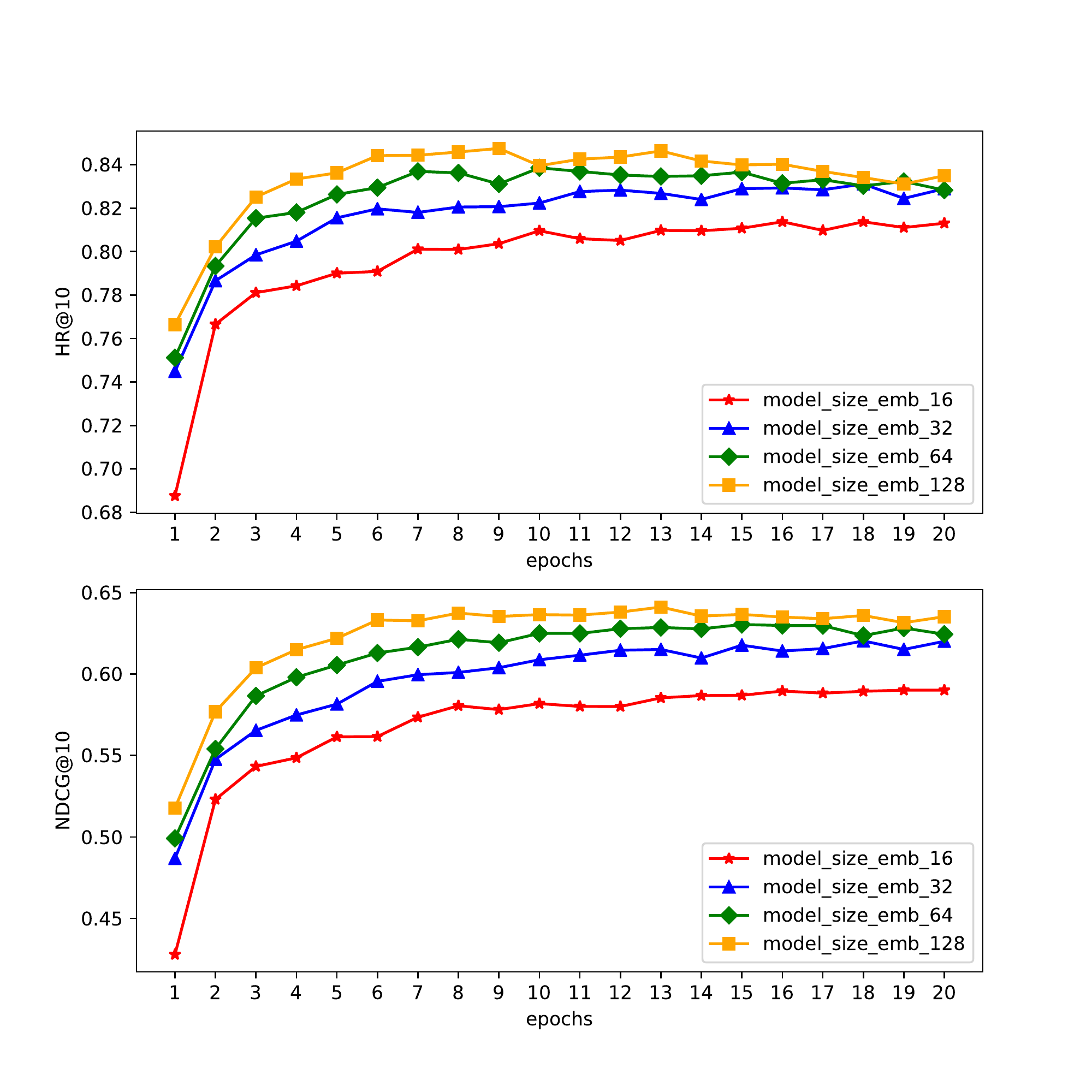}
	\caption{The performances of MPM $w.r.t$ different model size.}
	\label{fig: model_size}
\end{figure}

Considering the efficiency of model, it is curious to see whether employing a larger model size will be beneficial to the recommendation task. Towards this end, we further investigated MPM with different model size. To ensure the extraction of instant preferences, we keep the architecture of TCN unchanged. The other factors of model size are embedding size and the scale of the MLP module. We combine these two factors to determine the model size. When the embedding size is set to 16, the architecture of MLP is adopted to $[32,64,32,16]$, when the embedding size is 64, the MLP is set to $[128,256,128,64]$ and so on. We conduct experiments on ml-1m dataset to study the effects of model size. The experimental result is shown as figure \ref{fig: model_size}.

We can see that the larger the embedding size is, the better the performance of the model is. However, we cannot guarantee a larger model size will lead to better model performance, as overfitting could be a possible consequence. On the other hand, the improvement is not significant enough in both HR@10 and NDCG@10, as the model size grows up. This may cause by the fact that when model size grows larger, the gain from increasing model size can't trade off the complexity it brings in. As a consequence, we should tune the model size carefully for different datasets.

\section{Conclusions}


In this work, we explored the influence of unexpected behaviors, which is harmful to recommendations. We propose a multi-preferences model which can eliminate the effect of unexpected behaviors. The proposed model extracts instant preferences from recent historical interactions and detect which instant preferences are biased by unexpected behaviors under the guidance of the target item. Then, the output module eliminates the effects of unexpected behaviors based on the detection.  We also integrate a conventional latent factor model to learn the general preferences for recommendations. Extensive experiments are performed to show the effectiveness of our model.

In the future, we will extend our work in three directions. First, we attempt to mimic the interactions between users and items via graph neural networks to capture general preferences of users and characteristics of the items, since precise user and item embedding learning is the key to building a successful recommender system. Second, we plan to design a dedicated loss function instead of the general binary cross-entropy loss to get the most out of our model's performance.  In the end, we should also try some effect to bring in more auxiliary information for building a more robust recommender system.

\bibliographystyle{ieicetr}
\bibliography{multi_preferences_model}

~\\

\profile[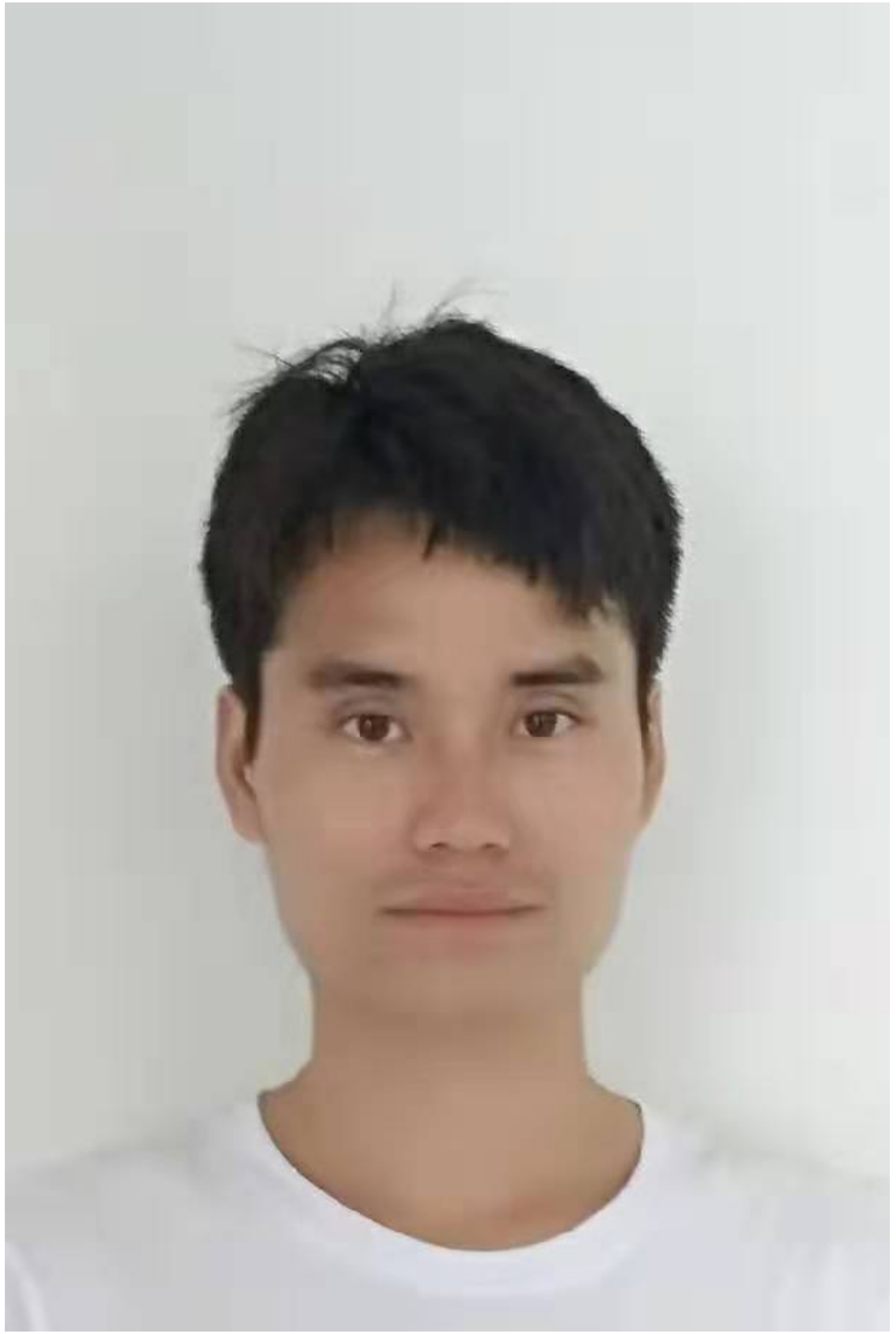]{Jie Chen (The First Author)}{born in 1995. MSc. His main research interests include Artificial Intelligence, Recommender System, Data Mining and Machine Learning.}
\profile[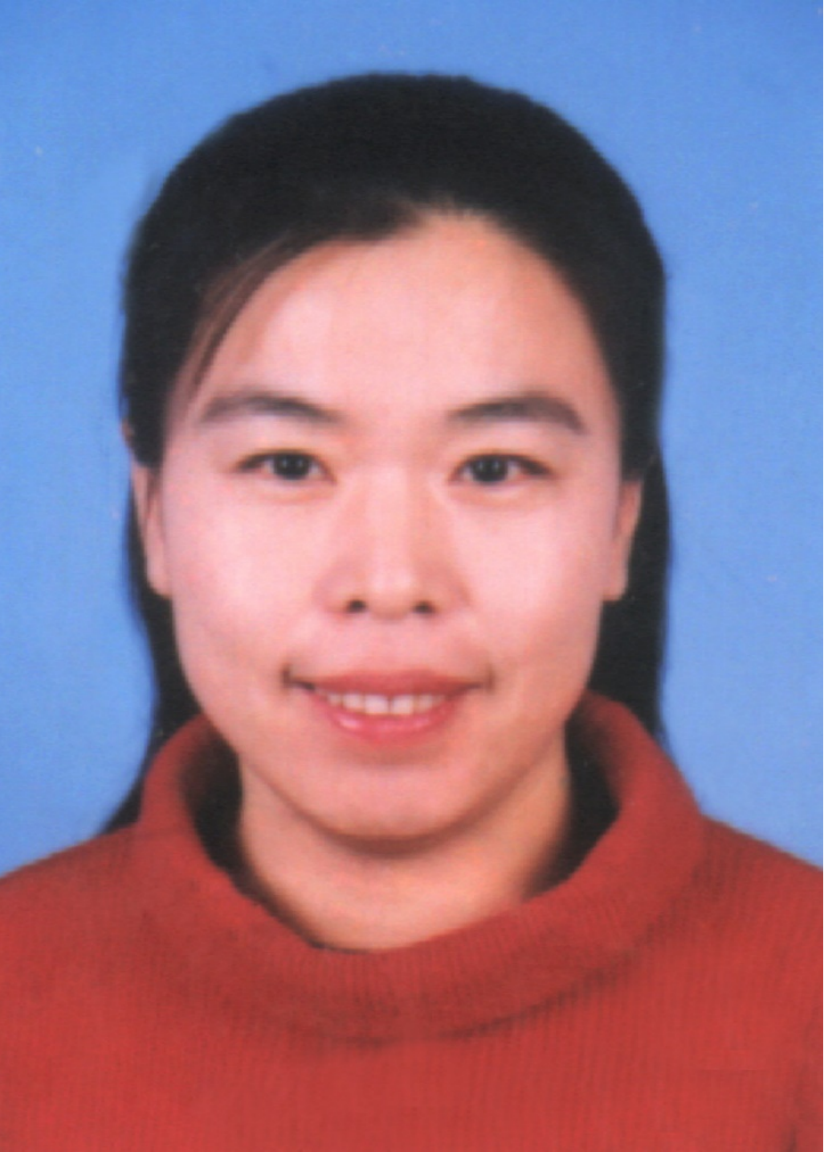]{LifenJiang(Corresponding Author)}{born in 1964. PhD, professor, PhD supervisor. Her main research interests include Artificial Intelligence, Pervasive Computing, and Machine Learning.}
\profile[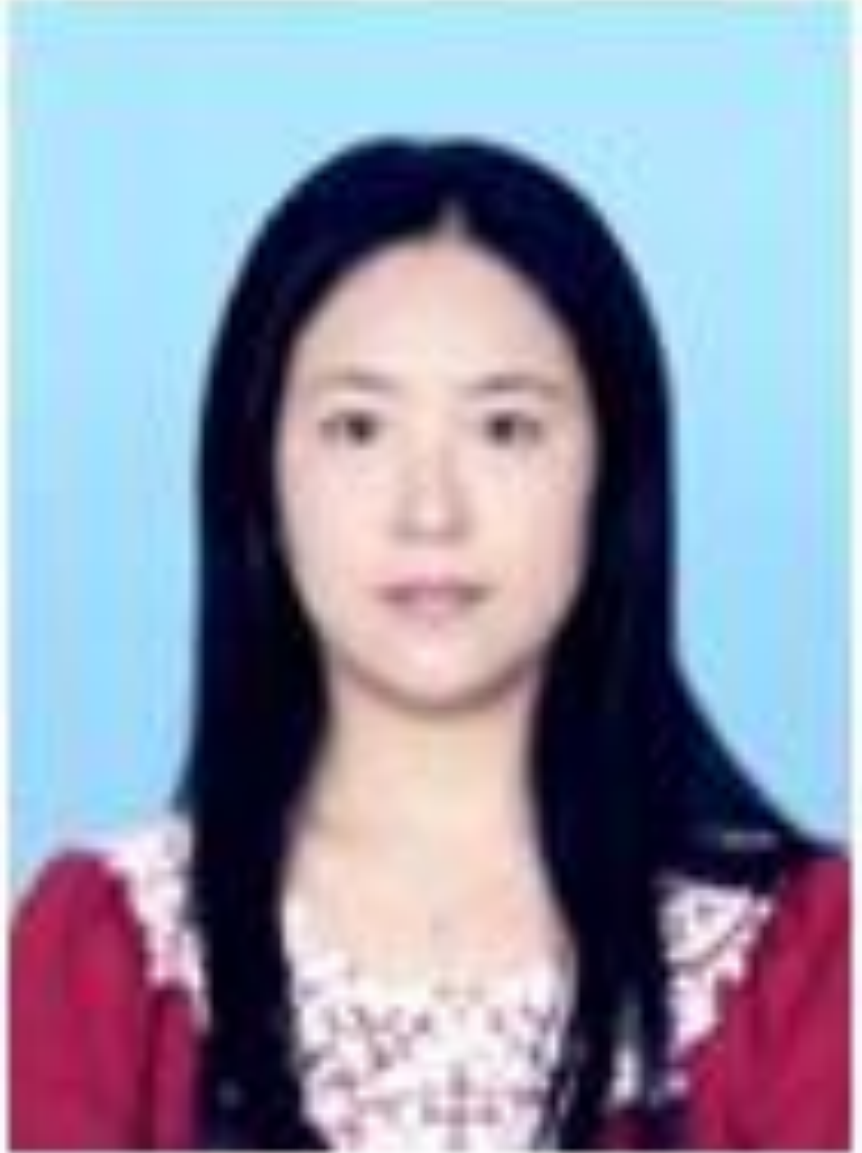]{Chunmei Ma}{born in 1985. PhD, lecturer. Her main research interests include Intelligent Transportation, Machine Learning, Mobile computing, Crowd Sensing , and Vehicular Ad Hoc Network.}
\profile[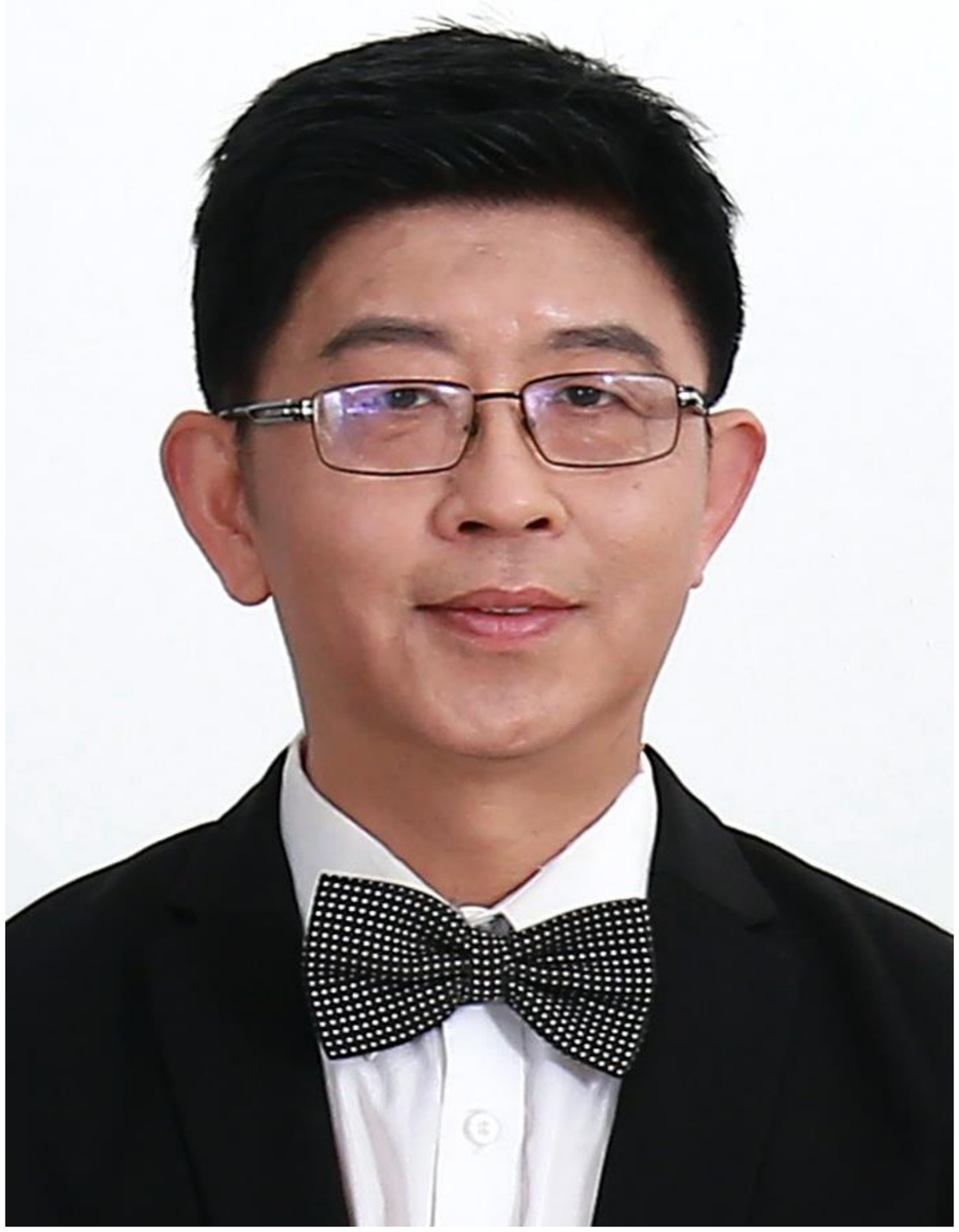]{Huazhi Sun}{born in 1961. PhD, professor, PhD supervisor. Dean of College of Computer and Information Engineering, Tianjin Normal University. His main research interests include Artificial Intelligence, Pattern Recognition, and Machine Learning.}

\end{document}